\documentclass{article}
\usepackage{ijcai20}

\usepackage{times}
\usepackage{soul}
\usepackage{url}
\usepackage[hidelinks]{hyperref}
\hypersetup{
    colorlinks=true,
    linkcolor=blue,
    filecolor=magenta,      
    urlcolor=cyan,
}
\usepackage[utf8]{inputenc}
\usepackage[small]{caption}
\usepackage{cite}
\usepackage{graphicx}
\usepackage{textcomp}
\usepackage{xcolor}
\usepackage{float}
\usepackage{enumerate}
\usepackage{caption}
\usepackage{subcaption}
\usepackage{graphicx}
\graphicspath{{figures/}}
\usepackage{mathtools,commath,amsmath,amssymb,amsfonts,bm}\interdisplaylinepenalty=2500
\usepackage{amsthm}
\usepackage{algpseudocode,algorithm}
\usepackage{tabularx,adjustbox}
\usepackage{todonotes}
\usepackage{breqn}
\clearpage
\extrafloats{100}
\RequirePackage{tcolorbox}
\tcbuselibrary{breakable}
\tcbset{parbox=false, left=1ex, right=1ex}

\usepackage{multirow}
\usepackage{xspace}
\usepackage{url}
\newcolumntype{C}[1]{>{\hsize=#1\hsize\centering\arraybackslash}X}

\def\vec#1{\mathchoice{\mbox{\boldmath$\displaystyle#1$}}
  {\mbox{\boldmath$\textstyle#1$}}
  {\mbox{\boldmath$\scriptstyle#1$}}
  {\mbox{\boldmath$\scriptscriptstyle#1$}}}

\newcommand{\etal}{\textit{et~al.}\xspace}
\newcommand{\ie}{\textit{i.e.},\xspace}

\newtheorem{definition}{Definition}

\algnewcommand\algorithmicinput{\textbf{Input:}}
\algnewcommand\Input{\item[\algorithmicinput]}
\algnewcommand\algorithmicoutput{\textbf{Output:}}
\algnewcommand\Output{\item[\algorithmicoutput]}
\algnewcommand\algorithmictier{\textbf{Role:}}
\algnewcommand\Tier{\item[\algorithmictier]}

\title{Collaborative Fairness in Federated Learning}

\author{
Lingjuan Lyu$^{1*}$
\and
Xinyi Xu$^{1*}$
\And
Qian Wang$^{3*}$
\affiliations
$^1$Department of Computer Science, National University of Singapore, Singapore\\
$^3$School of Cyber Science and Engineering, Wuhan University\\
\emails
Corresponding to: lyulj@comp.nus.edu.sg, xuxinyi@comp.nus.edu.sg, %yitongl4@student.unimelb.edu.au, 
qianwang@whu.edu.cn.
}

\begin{document}
\maketitle

\begin{abstract}
In current deep learning paradigms, local training or the \emph{Standalone} framework tends to result in overfitting and thus poor generalizability. This problem can be addressed by \emph{Distributed} or \emph{Federated Learning} (FL) that leverages a parameter server to aggregate model updates from individual participants. However, most existing Distributed or FL frameworks have overlooked an important aspect of participation: collaborative fairness. In particular, all participants can receive the same or similar models, regardless of their contributions. To address this issue, we investigate the collaborative fairness in FL, and propose a novel \emph{Collaborative Fair Federated Learning} (CFFL) framework which utilizes reputation to enforce participants to converge to different models, thus achieving fairness without compromising the predictive performance. Extensive experiments on benchmark datasets demonstrate that CFFL achieves high fairness, delivers comparable accuracy to the \textit{Distributed} framework, and outperforms the \textit{Standalone} framework. Our code is available on \href{https://github.com/XinyiYS/CollaborativeFairFederatedLearning}{github}.
\end{abstract}

% \begin{IEEEkeywords}
% Collaborative Learning, Fairness, Reputation.
% \end{IEEEkeywords}
\section{Introduction}
\label{sec:introduction}
Training complex deep neural networks on large-scale datasets is computationally expensive so it may not be feasible by a single participant. Moreover, training a complex model on a limited local dataset may lead to poor generalizability. In this context, Federated Learning (FL) emerged as a promising paradigm, as it provides a way for multiple devices/participants to jointly train a model whiling keeping their datasets local. The objective of FL is to derive a global model with better generalizability by leveraging the local datasets from multiple participants~\cite{mcmahan2017communication}.

However, most of the current FL paradigms~\cite{mcmahan2017communication,kairouz2019advances,yang2019federated,li2019federated} allow all participants to receive the same federated model in each communication round and in the end, regardless of their contributions. This is obviously unfair, because in practice not all participants contribute equally due to various reasons, such as the diverse quality/quantity of the data owned by different participants. Therefore, the data from some participants may lead to good model updates while updates from some other participants can even impair the model performance. Consider a motivating practical example: several banks may want to collaborate to build a credit score predictor for small and medium enterprises. However, larger banks with more data may be reluctant to train their local model based on high quality local data because sharing these high quality model parameters with smaller banks may potentially erode the market share of the larger bank~\cite{FL2019}. Furthermore, as the paradigm cannot distinguish the participants with high contributions from the ones with relatively low contributions, it is vulnerable to free-riders. Hence, %without collaborative fairness, the FL paradigm is faced with challenges in getting adopted in practice with participants from different organizations/companies/institutions. Consequently, this issue potentially hinders 
lacking collaborative fairness may hinder the formation and progress of a healthy FL ecosystem. Existing research on fairness mostly focuses on mitigating potential bias introduced to the model towards certain attributes~\cite{cummings2019compatibility,jagielski2018differentially}. The problem of treating the FL participants fairly according to their contributions remains open~\cite{FL2019}.

For any proposed solution or framework to be practical, it is essential to achieve fairness \emph{not} at the cost of model performance. In this work, we address this problem of treating FL participants fairly based on their contributions by proposing a \emph{Collaborative Fair Federated Learning} (CFFL) framework. Unlike existing work such as~\cite{Yu-et-al:2020AIES} which requires external monetary incentives for good behaviour, CFFL makes fundamental changes to the learning process in FL so that the participants will receive models with performance commensurate with their contributions, instead of the same FL model. CFFL achieves collaborative fairness with a reputation mechanism, which evaluates the contributions of the participants in the learning process and iteratively updates their respective reputations. We highlight the practical relevance of our CFFL in horizontally federated learning (HFL) to businesses (H2B)~\cite{lyu2020threats}, such as biomedical or financial institutions to whom collaborative fairness is very important.

Our work aims to achieve collaborative fairness in FL by adjusting the performance of the models allocated to each participant based on their contributions~\cite{lyu2020towards,lyu2020how}. Experiments on benchmark datasets demonstrate that CFFL achieves the highest fairness. %and in terms of predictive accuracy, performs comparably to the 
In terms of utility, the accuracy of the most contributive participant in CFFL is comparable to that of the \textit{Distributed} framework, and higher than that of the \textit{Standalone} framework. In the following sections, we interchangeably use Distributed/Federated.

%%%%%%%%%%%%%%%%%%%%%%%%%%%%%%%%%%%%%%%%
\section{Related Work}
\label{sec:Related_work}
In this section, we review the relevant literature on fairness in FL to position our research in relation to existing research.

One existing approach for promoting collaborative fairness among federated participants is based on incentive schemes~\cite{yang2019federated}. In principle, participants shall receive payoffs commensurate with their contributions. Equal division is an example of egalitarian profit-sharing \cite{Yang-et-al:2017IEEE}. Under this scheme, the available total payoff at a given round is equally divided among all participants. Under the Individual profit-sharing scheme \cite{Yang-et-al:2017IEEE}, each participant $i$'s own contribution to the collective (assuming a singleton collective of $i$) is used to determine his share of the total payoff.

The Labour Union game \cite{Gollapudi-et-al:2017} profit-sharing scheme determines a participant's share of the total payoff based on his marginal contribution to the utility of the collective formed by his predecessors. %(i.e. each participant's marginal contribution is computed based on the same sequence as they joined the federation). 
The Fair-value game scheme \cite{Gollapudi-et-al:2017} is a marginal loss-based scheme. Under this scheme, a participant's share of the total payoff is determined according to the sequence of the participants leaving the collective. The Shapley game profit-sharing scheme \cite{Gollapudi-et-al:2017} is also a marginal contribution-based scheme. Unlike the Labour Union game, Shapley game aims to eliminate the effect of the sequence in which the participants join the collective in order to obtain a fairer estimate of their marginal contributions to the collective. %They achieve so by computing the average of the marginal contributions for each participant under all different permutations of them joining the collective relative to other participants. As a result, 
However, the complexity of this approach is exponential in the number of participants, making it prohibitively expensive for large-scaled FL in practice.

For gradient-based FL approaches, the gradient information can be regarded as a useful source of data. However, in these cases, output agreement-based rewards are hard to apply as mutual information requires a multi-task setting which is usually not present in such cases. Thus, among these three categories of schemes, model accuracy is the most relevant way of designing rewards for FL. There are two emerging federated learning incentive schemes focused on model improvement.

Richardson \etal~\cite{richardson2019rewarding} proposed a scheme which pays for marginal improvements brought about by model updates. The sum of improvements might result in overestimation of contribution. Thus, the proposed approach also includes a model for correcting the overestimation issue. This scheme ensures that payment is proportional to model quality improvement, which means the budget for achieving a target model quality level is predictable. It also ensures that data owners who submit model updates early receive a higher reward. This in turn motivates them to participate even in early stages of the federated model training process. 

Yu \etal~\cite{Yu-et-al:2020AIES} proposed a joint objective optimization-based approach that in addition to the contributions of the participants, takes costs and waiting time into account in order to achieve additional notions of fairness when distributing payoffs to the FL participants. %Different from the aforementioned approaches, our proposed framework provides an alternative paradigm, in which each participant is allocated with a different version of the FL model with performance commensurate with the contributions.%does not utilize monetary payoffs to achieve fair treatment of FL participants. Instead, it allocates each of them a slightly different version of the FL model with performance commensurate with the contributions. This introduces an alternative paradigm to existing FL paradigm where all participants receive the same final FL model.

%%%%%%%%%%%%%%%%%%%%%%%%%%%%%%%%%%%%%%%%
\section{The CFFL Framework}
\label{sec:CFFL}
\subsection{Collaborative Fairness}

%Collaborative fairness enforces participants to converge to different final models based on their contributions, hence the notion of fairness most relevant to our purpose is \textit{Fairness through Awareness}.
Different from the existing approaches, our proposed framework provides an alternative paradigm, in which participants are allocated with different versions of the FL model with performance commensurate with their contributions.
Under this context, we define collaborative fairness as follows.

\begin{definition}\label{def:fairness}
Collaborative fairness. In a federated system, a high-contribution participant should be rewarded with a better performing local model than a low-contribution participant. Mathematically, fairness can be %quantifiably represented 
quantified by the correlation coefficient between the contributions of participants and their respective final model accuracies. 
\end{definition}

\subsection{Fairness via Reputation}
In our CFFL, we modify FL by allowing participants to download only the \emph{allocated aggregated updates according to their reputations}. The server manages a reputation list for all participants, and updates it according to the quality of the uploaded gradients of each participant in each communication round. The upload rate -- $\theta_u$ -- denotes the proportion of parameters of which gradients are uploaded, i.e., if $\theta_u=1$, gradients of all parameters are uploaded; if $\theta_u=0.1$, gradients of only 10\% the parameters are uploaded. We further denote the selected set of gradients as $S$, corresponding to $\theta_u$ gradients selected according to the ``\textit{largest values}" criterion: sort the gradients in $\Delta \vec{w}_j$ (by their %moduli
magnitude), and upload $\theta_u$ of them, starting from the largest. %select the largest $\theta_u$ of them. 
Specifically, the server separately evaluates the validation accuracy of participant $j$ by integrating $j$'s uploaded gradients. In particular, if $\theta_u=1$, $\Delta (\vec{w}_j)^S \triangleq \Delta \vec{w}_j$, the server can derive participant $j$'s entire model $\vec{w}_j$, as all participants are initialized with the same parameters in the beginning. The server then computes the validation accuracy of participant $j$ based on $\vec{w}_j$ as $vacc_j \gets V(\vec{w}_j + \Delta (\vec{w}_j)^S)$, here $V$ denotes the validation dataset. If $\theta_u \neq 1$, the server simply integrates participant $j$'s uploaded gradients $\Delta (\vec{w}_j)^S$ into an auxiliary model $\vec{w}_g$ kept by the server to compute participant $j$'s validation accuracy as $vacc_j \gets V(\vec{w}_g + \Delta (\vec{w}_j)^S)$. Note $\vec{w}_g$ is an auxiliary model maintained by the server to %collect, 
aggregate gradients and calculate participants' reputations, and its parameters are \emph{not} broadcast to individual participants as in the standard FL systems.

Then the server normalizes $vacc_j$ and passes the normalized $vacc_j$ through a $sinh(\alpha)$ function in Eq.~\eqref{eq:reputation} to calculate the reputation $c_j$ of participant $j$ in each communication round.
\begin{equation}\label{eq:reputation}
	c_j=sinh(\alpha*x)
\end{equation}
Here $x$ is the normalized $vacc_j$, so the higher $x$, the more informative participant $j$'s uploaded gradients are. $sinh(\alpha)$ is introduced as a \textit{punishment function}, and $\alpha$ denotes the \textit{punishment factor}, used to distinguish the reputations among participants based on how informative their uploaded gradients are. The server iteratively updates the reputation of each participant separately based on the calculated reputation in each round and its historical reputation. The high-contribution participant will be highly rated by the server, while the low-contribution participant can be detected and even isolated from the federated system, avoiding the low-contribution participants from dominating the whole system, or free-riding. 

This computed reputation determines the number of aggregated gradients each participant will be allocated in the subsequent communication round. The higher the reputation of participant $j$, the more aggregated gradients will be allocated to participant $j$. The aggregated gradients refer to the collection of gradients from all participants, %through some aggregation, such as FedAvg~\cite{mcmahan2017communication}, 
and are used here as a form of reward in each communication round. The detailed realization of CFFL is in Algorithm~\ref{Algorithm:Fair_FL}. In each communication round, each participant uploads $\theta_u$ fraction of clipped gradients to the server, and server updates the reputation based on the performance of these uploaded gradients on a validation set, and determines the number of aggregated updates to allocate to each participant. We adopt gradient clipping to reduce the impact of noise from abnormal examples/outliers.

\begin{algorithm}[!htbp]
%\setstretch{1.35}
\caption{Collaborative Fair Federated Learning}\label{Algorithm:Fair_FL}
\small
\begin{algorithmic} 
%\State \textbf{Input: $C$, $\vec{w^g}$, $\Delta \vec{w}_j$, $\lambda_j$, $\vec{w}_i$, $V$}
%\State \textbf{Output: updated parameters $\vec{w}_i'$, reputation ${c_i}'$}
\Require %Communication rounds $C$, 
reputable participant set $R$; auxiliary model $\vec{w_g}$ kept by server; local model $\vec{w}_j$; local model updates $\Delta \vec{w}_j$; upload rate $\theta_u$; validation set $V$; local epochs $E$; $c_j^o$: reputation of previous round; $D_j$: data owned by each participant; %$n_j$: number of examples owned by each participant; 
data shard vector $n=\{n_1,\cdots,n_{|R|}\}$; %$class_j$: number of classes owned by each participant; 
class shard vector $class=\{class_1,\cdots,class_{|R|}\}$.%$n \in \mathbb{Z}^m$ .%$n \in \mathbb{Z}^m$ .%sharing level $\lambda_j$, 
%\State At current epoch, suppose participant $i$ aims to download total $d_i$ gradients from all participants in $C$, while participant $j \in C \setminus i$ can at most upload $\lambda_j \times |\Delta \vec{w}_j|$ gradients. Server updates the reputation list based on the gradients of participant $j \in C$ as follows:
%\State In each communication round, each participant sends $\theta_u$ fraction of gradients to the server, and server updates its reputation as per its performance on the public validation set, then determines how many aggregated updates this participant can download. Each participant is initialized with the same parameter to start with.

%\For{$c \in C$}
     \Tier participant $j$
     \If{$j \in R$}
%     \State Downloads the allocated global parameter $\vec{w_g^j}$ from server and replaces the corresponding elements in local model $\vec{w}_j$ with $\vec{w_g^j}$;

     \State Runs SGD on local data by using current local model $\vec{w}_j$ and computes gradient vector: $\Delta \vec{w}_j  \gets \texttt{SGD}(\vec{w}_j, D_j)$ 
     
%     updates the previous local parameters $\vec{w}_j$ as $\vec{w}_j'$;
%     \State  Computes gradient vector $\Delta \vec{w}_j=\vec{w}_j'-\vec{w}_j$;
    %  Conducts local training to compute gradients: 
    %\State $d_j^i=min(c_i^j*d_i, \lambda_j*|\Delta \vec{w}_j|)$
    \State Clips gradient vector:  $\Delta \vec{w}_j \gets clip(\Delta \vec{w}_j)$
    \State Sends the selected gradients $\Delta (\vec{w}_j)^S$ of size $\theta_u*|\Delta \vec{w}_j|$ to the server, %$\Delta 
    %$(\vec{w}_j)^S$ are grouped into set $S$, 
    according to the ``largest values" criterion;
    %\textbf{largest values}: gradients of $\Delta \vec{w}_j$ 
    %sort gradients in $\Delta \vec{w}_j$, and upload $\theta_u$ of them, starting from the largest;
    \State Downloads the allocated %aggregated 
    updates %$\Delta \vec{w_g^j}- \frac{n_j}{max(n)} \Delta (\vec{w}_j)^S$ or $\Delta \vec{w_g^j}- \frac{class_j}{max(class)} \Delta (\vec{w}_j)^S$ 
    from the server, %excludes a weighted version of its own uploaded updates, %its own uploaded updates to derive the other participants' aggregated updates, %: $\Delta \vec{w_g^j}-\Delta (\vec{w}_j)^S$; 
    %\State  
    which is then integrated with all its local updates %and the other participants' aggregated updates to update local model
    as: $\vec{w}_j' \gets \vec{w}_j+\Delta \vec{w}_j+\Delta \vec{w_g^j}- \frac{n_j}{max(n)} \Delta (\vec{w}_j)^S$ (imbalanced data size) or $\vec{w}_j' \gets \vec{w}_j+\Delta \vec{w}_j+\Delta \vec{w_g^j}- \frac{class_j}{max(class)} \Delta (\vec{w}_j)^S$ (imbalanced class number).  
    \EndIf
    \vspace{2mm}
    
    \Tier Server
    \\\textbf{Updates aggregation}:
    \If{data size is imbalanced}
    \State $\Delta \vec{w}_g \gets {\textstyle\sum}_{j \in R} \Delta (\vec{w}_j)^S \times \frac{ n_j}{sum(n)}$. 
    \EndIf
    
    \If{class number is imbalanced}
    \State $\Delta \vec{w}_g \gets {\textstyle\sum}_{j \in R} \Delta (\vec{w}_j)^S \times \frac{ class_j}{max(class)}$. 
    \EndIf
    % \State \textbf{Updates aggregation}: $\Delta \vec{w}_g \gets {\textstyle\sum}_{j \in R} \Delta (\vec{w}_j)^S \times \frac{n_j}{sum(n)}$. 
    % \For{$j \in R$}
    % \If{$\theta_u=1$ and $E=1$}
    \If{$\theta_u=1$}
    \For{$j \in R$}
    	\State $vacc_j \gets V(\vec{w}_j+\Delta (\vec{w}_j)^S)$.
	\State Updates local model of participant $j$ kept by the server: $\vec{w}_j' \leftarrow \vec{w}_j+\Delta (\vec{w}_j)^S$ for next round of reputation evaluation.
	\EndFor
%    \EndIf
%    \If{$\theta_u \neq 1$}
   \Else	
    \For{$j \in R$}
    	\State $vacc_j \gets V(\vec{w}_g+\Delta (\vec{w}_j)^S)$.
    \EndFor
    \State Updates temp model maintained by server $\vec{w}_g'=\vec{w}_g +  \Delta \vec{w}_g$ for next round of reputation evaluation.
    \EndIf
    % \EndFor
%    \If{$\theta_u \neq 1$}
%     \State Updates global parameter $\vec{w}_g'=\vec{w}_g+{\textstyle\sum}_{j \in R} \Delta (\vec{w}_j)^S$ for next round of reputation evaluation.
%    \EndIf
    \For{$j \in R$}
 	\State $c_j \gets sinh(\alpha*\frac{vacc_j}{{\textstyle\sum}_{j \in R} vacc_j})$, $c_j' \gets c_j^o*0.5+c_j*0.5$ 
    % \State $c_j' \gets c_j^o*0.2+c_j*0.8$ , where $c_j^o$ refers to the past reputation %in the previous communication round.
    \EndFor
    % \State Integrates historical reputation: $c_j'=c_j^o*0.2+c_j*0.8$, where $c_j^o$ refers to the reputation in the previous communication round.
    \State \textbf{Reputation normalisation}: 
      $c_j' \gets \frac{c_j'}{{\textstyle\sum}_{j \in R} c_j'}$
    \If{${c_j}'<c_{th}$}
    \State $R \gets R\setminus \{j\} $, repeat reputation normalisation.
    %Server flags participant $j$ as "low-contribution" and removes it from $R$,
    \EndIf
    % \State \textbf{Normalise reputations} if necessary

    \For{$j \in R$} 
    \If{data size is imbalanced}
    \State $num_j \gets \frac{c_j'}{max(c)} * \frac{n_j}{max(n)} *|\Delta \vec{w}_g| $
    \EndIf
    
    \If{class number is imbalanced}
    \State $num_j \gets \frac{c_j'}{max(c)} * \frac{class_j}{max(class)} *|\Delta \vec{w}_g| $
    \EndIf
    \State Groups $num_j$ aggregated updates into $\Delta \vec{w_g^j}$ according to the ``largest values" criterion, and allocates an adjusted version $\Delta \vec{w_g^j}- \frac{n_j}{max(n)} \Delta (\vec{w}_j)^S$ (imbalanced data size) or $\Delta \vec{w_g^j}- \frac{class_j}{max(class)} \Delta (\vec{w}_j)^S$ (imbalanced class number) to participant $j$.
    \EndFor
\end{algorithmic}
\end{algorithm}

\subsection{Quantification of Fairness}
In this work, we quantify collaborative fairness via the correlation coefficient between participant contributions (X-axis: test accuracies of standalone models which characterize their individual learning capabilities on their own local datasets) and participant rewards (Y-axis: test accuracies of final models received by the participants). 

Participants with higher standalone accuracies empirically contribute more.
Therefore, the X-axis can be expressed by Equation~\ref{eq:x_axis}, where $sacc_j$ denotes the standalone model accuracy of participant $j$:
\begin{equation}\label{eq:x_axis}
\vec{x}=\{sacc_1,\cdots,sacc_n\}
\end{equation}

Similarly, Y-axis can be expressed by Equation~\ref{eq:y_axis}, where $acc_j$ represents the final model accuracy of participant $j$:
\begin{equation}\label{eq:y_axis}
\vec{y}=\{acc_1,\cdots,acc_n\}
\end{equation}

As the Y-axis measures the respective model performance of different participants after collaboration, it is expected to be positively correlated with the X-axis for a good measure of fairness. Hence, we formally quantify collaborative fairness in Equation~\ref{eq:fairness}:
\begin{equation}\label{eq:fairness}
r_{xy}=\frac{{\textstyle\sum}_{i=1}^n (x_i-\bar{x})(y_i-\bar{y})}{(n-1)s_xs_y}
\end{equation}
where $\bar{x}$ and $\bar{y}$ are the sample means of $\vec{x}$ and $\vec{y}$, $s_x$ and $s_y$ are the corrected standard deviations. The range of fairness falls within [-1,1], with higher values implying good fairness. Conversely, negative coefficient implies poor fairness. 

%%%%%%%%%%%%%%%%%%%%%%%%%%%%%%%%%%%%%%%%
\section{Experimental Evaluation}
\label{sec:Performance}
\subsection{Datasets}
We implement experiments on two benchmark datasets. The first is the MNIST dataset\footnote{\url{http://yann.lecun.com/exdb/mnist/}} for handwritten digit recognition, consisting of 60,000 training examples and 10,000 test examples. The second is the Adult Census dataset\footnote{\url{http://archive.ics.uci.edu/ml/datasets/Adult}}. %, which contains 14 attributes, including age, race, education level, marital status, occupation, etc. 
This dataset is commonly used to predict whether an individual makes over 50K dollars in a year (binary). There are total 48,843 records, %with 24\% (11687) records over 50,000, while the remaining 76\% (37155) under 50,000. 
we manually balance the dataset to have 11687 records over 50K and 11687 records under 50K, resulting in total 23374 records. We then conduct an 80-20 train-test split. For all datasets, we randomly choose 10\% of training examples as the validation set. 

\subsection{Baselines}
We demonstrate the effectiveness of our proposed CFFL framework through comparison with the following frameworks.%\textit{Standalone} framework and two representative \textit{Distributed} baselines.%: FedAvg~\cite{mcmahan2017communication} and DSSGD~\cite{shokri2015privacy}.% for FL. 

\textit{Standalone} framework allows participants to train standalone models on local datasets without collaboration. This framework delivers minimum utility, because each participant is susceptible to falling into local optima when training alone. In addition, we remark that there is no concrete concept of collaborative fairness in the \textit{Standalone} framework, because participants do not collaborate.

\textit{Distributed} framework enables participants to train independently and concurrently, and share their gradients or model parameters to achieve a better global model. For comparison, we choose two representative \textit{Distributed} baselines, including FedAvg~\cite{mcmahan2017communication} and DSSGD~\cite{shokri2015privacy}.%We implement two commonly adopted frameworks for distributed/federated learning. We follow~\cite{mcmahan2017communication} to compare with the standard approach of Federated Averaging (FedAvg) in FL. In addition, as demonstrated in~\cite{shokri2015privacy}, Distributed Selective SGD (DSSGD) can achieve even higher accuracy than the centralized SGD because updating only a small fraction of parameters in each round regularizes the training by the neural network weights from jointly ``remembering'' the training data, and helps avoids overfitting. Hence, we consider DSSGD for the analysis of the \textit{Distributed} framework and omit the centralized framework. 

Furthermore, we investigate different upload rates $\theta_u=0.1$ and $\theta_u=1$~\cite{shokri2015privacy}, where gradients are uploaded according to the ``largest values'' criterion when $\theta_u=0.1$. The rationale behind introducing an upload rate less than 1 is to reduce overfitting and to save communication overhead. %We observe a similar practice by the FedAvg algorithm that in a large-scale FL system, it may randomly select a subset of participants in each communication round for parameter update and sharing, in order to reduce the risk of overfitting and to reduce communication cost.

\subsection{Experimental Setup}
\label{sec:Setup}
Due to the fact that the data are often heterogeneous across participants both in terms of size and distribution, we investigate the following two realistic scenarios:

\textbf{Imbalanced data size.} To simulate data size heterogeneity, we follow a power law to randomly partition total \{3000,6000,12000\} MNIST examples among \{5,10,20\} participants respectively. Similarly, for Adult dataset, \{4000,8000,12000\} examples are randomly partitioned among \{5,10,20\} participants. In this way, each participant has a distinctly different number of examples, with the first participant has the least and the last participant has the most. We remark that the purpose of allocating on average $600$ MNIST examples for each participant is to fairly compare with Shokri \etal~\cite{shokri2015privacy}, where each participant has a small number of $600$ local examples to simulate data scarcity which necessitates collaboration. 

\textbf{Imbalanced class numbers}. 
To examine data distribution heterogeneity, we vary the class numbers in the data of each participant, increasing from the first participant to the last. For this scenario, we only investigate the MNIST dataset. We distribute classes in a linspace manner, for example, participant-$\{1,2,3,4,5\}$ own \{1,3,5,7,10\} classes from MNIST dataset respectively. In more detail, for MNIST with total 10 classes and 5 participants, we simulate the first participant has data from only $1$ class, while the last participant has data from $10$ classes. We first partition the training dataset according to the labels, and then sample and assign subsets of training data with corresponding labels to the participants. Note that each participant still has the same number of examples, \ie 600 examples. 

\textbf{Model and Hyper-Parameters}. For MNIST \textit{Imbalanced data size} experiment, we use a two-layer fully connected neural network with 128 and 64 units respectively. The hyperparameters are: local epochs $E=2$, local batch size $B=16$, and local learning rate $lr=0.15$ for $P=5$ and $lr=0.25$ for $P=\{10, 20\}$, with exponential decay of $\gamma = 0.977$, gradient clipping between $[-0.01, 0.01]$, with a total of 30 communication rounds. For MNIST \textit{Imbalanced class numbers} experiment, the same neural network architecture is used. The hyperparameters are: local epochs $E=1$, local batch size $B=16$, and local learning rate $lr=0.15$ for $P=\{5, 10, 20\}$, with exponential decay of $\gamma = 0.977$, gradient clipping between $[-0.01, 0.01]$, with a total of 50 communication rounds. For Adult, we use a single layer fully connected neural network with 32 units. The hyperparameters are: local epochs $E=2$, local batch size $B=16$, and local learning rate $lr=0.03$ with exponential decay of $\gamma = 0.977$, gradient clipping between $[-0.01, 0.01]$, with a total of 30 communication rounds. 

Furthermore, to reduce the impact of different initializations and avoid non-convergence, we initialize the same model parameter $\vec{w}_0$ for all participants and the server $\vec{w}_g$. %Subsequently, participants conduct local training on their individual training sets and update their respective model parameters $\vec{w}_j$. 
For all the experiments, we empirically set the reputation threshold via grid search as follows:
$c_{th}=\frac{1}{|R|}*\frac{1}{3}$ for imbalanced data size, and $c_{th}=\frac{1}{|R|}*\frac{1}{6}$ for imbalanced class numbers, where $|R|$ is the number of participants with reputations higher than the threshold. For the punishment factor, we empirically choose $\alpha=5$. Stochastic Gradient Descent(SGD) is used as the optimization technique throughout.

\textbf{Communication Protocol}. In standard FL, one global model is given to all participants, both during and at the end of the training process. Such a setup forbids the calculation of our definition of fairness via the pearson coefficient, when all participants have the same reward. To mitigate this, we follow~\cite{shokri2015privacy} to adopt the round-robin communication protocol for DSSGD and FedAvg. %the \textit{Distributed} framework. 
In each communication round, participants upload \textit{parameter updates} and download \textit{parameters} in sequence, leading to models with insignificant performance differences, so that their test accuracies can be used for the calculation of fairness. %For this same reason, we did \textit{not} explicitly modify the standard FedAvg algorithm to accommodate the calculation of fairness, and only considered FedAvg as a performance comparison baseline.

\subsection{Experimental Results}
\label{sec:results}

\begin{table*}[ht]
\caption{Fairness [\%] of DSSGD, FedAvg and CFFL under varying participant number settings (P-$k$), pretraining status and upload rate $\theta_u$. %\emph{FedAvg} not included as our fairness is not well-defined for it
.}% (P-$k$ indicates there are $k$ participants in the experiments).}
\label{tbl:MNIST_Adult_fairness}
\centering
%\scalebox{0.75}{\begin{tabularx}{\linewidth}{|p{0.45cm}|c|c|c|c|c|c|c|c|}
\resizebox*{1.0\textwidth}{!}{
\begin{tabular}{|c|c|c|c|c|c|c|c|c|c|c|c|c|c|c|}
\hline
%\multirow{3}{*}{} 
Dataset & \multicolumn{7}{c|}{MNIST} & \multicolumn{7}{c|}{Adult}
\tabularnewline
\hline
%\cline{2-9}
Framework & \multicolumn{1}{c|}{FedAvg}& \multicolumn{2}{c|}{DSSGD} & \multicolumn{4}{c|}{CFFL} & \multicolumn{1}{c|}{FedAvg}& \multicolumn{2}{c|}{DSSGD} & \multicolumn{4}{c|}{CFFL}
\tabularnewline
\hline
Pretrain & \multicolumn{1}{c|}{NA} & \multicolumn{2}{c|}{NA} & \multicolumn{2}{c|}{1} & \multicolumn{2}{c|}{0} & \multicolumn{1}{c|}{NA} & \multicolumn{2}{c|}{NA} & \multicolumn{2}{c|}{1} & \multicolumn{2}{c|}{0} 
\tabularnewline
\hline
$\theta_u$ & NA& $0.1$ & $1$ & $0.1$ & $1$ & $0.1$ & $1$ & NA& $0.1$ & $1$ & $0.1$ & $1$ & $0.1$ & $1$
\tabularnewline
\hline
\textit{P5}  
            & 3.08  & 90.72 & 84.61 & 99.63 & 98.66 & \textbf{99.76} & 99.02 %MNIST
            & -3.33 & 15.61 & 35.71 & 98.50 & 97.75 & 98.44 & \textbf{99.37} %ADULT
\tabularnewline
\hline
\textit{P10} 
            & -50.47 & -78.18 & 90.67 & 97.90 & 97.30 & 98.55 & \textbf{98.74}
            & 44.27 & 62.30 & 56.60 & 88.00 & \textbf{93.07} & 92.00 & 91.95
\tabularnewline
\hline
\textit{P20} 
            & 60.41 & -81.77 & 80.45 & \textbf{99.23} & 96.28 & 98.52 & 98.51
            & -34.32 & 60.30 & 58.01 & \textbf{84.41} & 82.46 & 80.56 & 79.52
\tabularnewline
\hline
\end{tabular}}
\end{table*}

\textbf{Fairness comparison}.
Table~\ref{tbl:MNIST_Adult_fairness} lists the calculated fairness of %the \textit{Distributed} framework using 
DSSGD, FedAvg and CFFL over MNIST and Adult under varying participant number settings from $\{5,10,20\}$, different pretraining status from $\{1,0\}$, and different upload rates $\theta_u$ from $\{0.1, 1\}$. From the high values of fairness (some close to the theoretical limit of 1.0), we conclude that CFFL achieves good fairness, confirming the intuition behind our notion of fairness: the participants with higher contributions are rewarded with better-performing models. Moreover, pretrain=1 can lead to slightly higher fairness than pretrain=0. This is attributed to the individual pretraining of 5 epochs before collaborative learning starts, because the participants' models have already moved towards their respective model optimum. Note that pretraining is \textit{only} conducted for CFFL. We also observe that DSSGD and FedAvg yield significantly lower fairness than our CFFL. This is expected since neither the communication protocol nor the learning algorithm incorporates the concept of fairness.

\begin{table}[!htp]
\caption{Maximum 
Accuracy [\%] over MNIST and Adult of varying participant number settings, achieved by DSSGD, FedAvg, \emph{Standalone} framework, and our CFFL ($\theta_u=0.1$, where CFFL* denotes CFFL with pretraining).
}
\label{tbl:MNIST_Adult_acc}
\centering
\begin{tabularx}{\linewidth}{|l|*{6}{C{1}|}}
\hline
\multirow{2}{*}{Framework} & \multicolumn{3}{c|}{MNIST} & \multicolumn{3}{c|}{Adult}
\tabularnewline
\cline{2-7}
 & P5 & P10 & P20 & P5 & P10 & P20 
\tabularnewline
\hline
\textit{DSSGD}
& 93.28 & 94.20 & 82.36
& 81.94 & 82.78 & 82.07
\tabularnewline
\hline
\textit{FedAvg}
& \textbf{93.62} & \textbf{95.32} & \textbf{96.26}
& \textbf{82.58} & \textbf{83.14} & \textbf{83.16}
\tabularnewline
\hline
\textit{Standalone} 
& 90.30 & 90.88 & 90.64
& 81.93 & 82.31 & 82.07
\tabularnewline
\hline
\textit{CFFL} 
& 91.83 & 93.00 & 93.25
& 81.96 & 82.63 & 82.72 
\tabularnewline
\hline
\textit{CFFL*} 
& 91.85 & 92.85 & 93.34
& 81.89 & 82.63 & 82.63
\tabularnewline
\hline
\end{tabularx}
\end{table}

\textbf{Accuracy comparison}. 
Table~\ref{tbl:MNIST_Adult_acc} reports the corresponding accuracies on MNIST and Adult datasets of $\{5,10,20\}$ participants when $\theta_u=0.1$. Here we report the best accuracy achieved among the participants, because CFFL enables participants to converge to different final models, so we expect the most contributive participant to receive a model with the highest accuracy comparable to both \textit{Distributed} frameworks. For the \textit{Standalone} framework, we show the accuracy of the same participant. It can be observed that CFFL obtains comparable accuracy to DSSGD and FedAvg, and always attains higher accuracy than the \textit{Standalone} framework. For example, for MNIST with 20 participants, our CFFL (CFFL*) achieves 93.25 (93.34)\% test accuracy, higher than the \textit{Standalone} framework (90.64\%), and slightly lower than FedAvg (96.26\%). The observation that DSSGD achieves lowest accuracy in this setting can be attributed to its higher instability and fluctuations during training. %The fairness results in Table~\ref{tbl:MNIST_Adult_fairness}, and the accuracy results in Table~\ref{tbl:MNIST_Adult_acc} demonstrate that our proposed CFFL achieves \emph{reasonable fairness without significantly compromising accuracy} under various settings.

\begin{figure*}[!t]
%\resizebox{.5\totalheight}{!}{
\centering
        \begin{subfigure}[ht]{0.19\textwidth}
                \includegraphics[width=3.0cm,height=2.9cm]{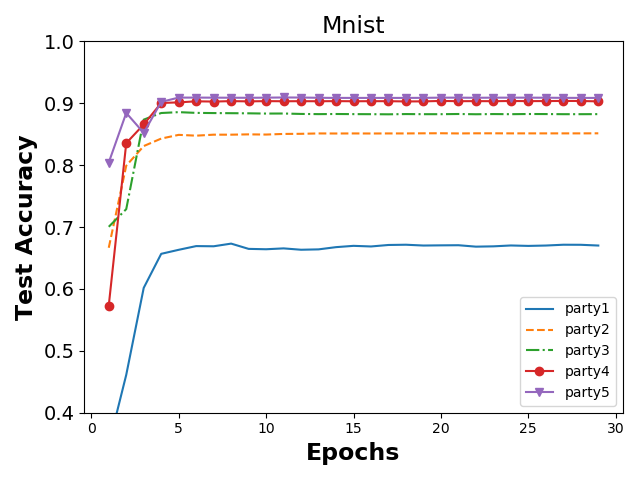}
        \end{subfigure}
        \begin{subfigure}[ht]{0.19\textwidth}
                \includegraphics[width=3.0cm,height=2.9cm]{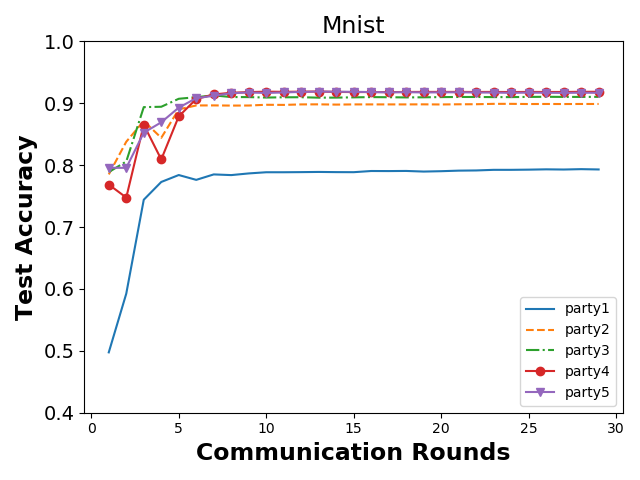}
        \end{subfigure}
         \begin{subfigure}[ht]{0.19\textwidth}
                 \includegraphics[width=3.0cm,height=2.9cm]{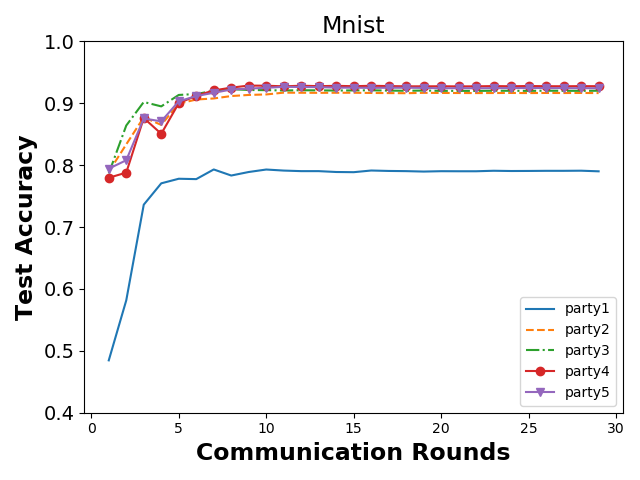}
        \end{subfigure}
        \begin{subfigure}[ht]{0.19\textwidth}
                \includegraphics[width=3.0cm,height=2.9cm]{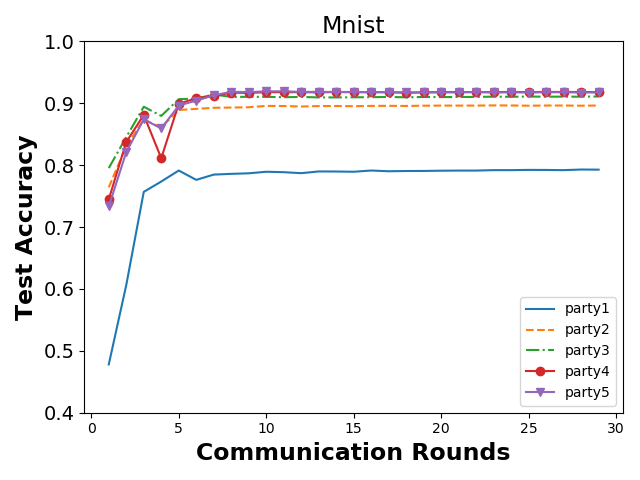}
        \end{subfigure}
         \begin{subfigure}[ht]{0.19\textwidth}
                 \includegraphics[width=3.0cm,height=2.9cm]{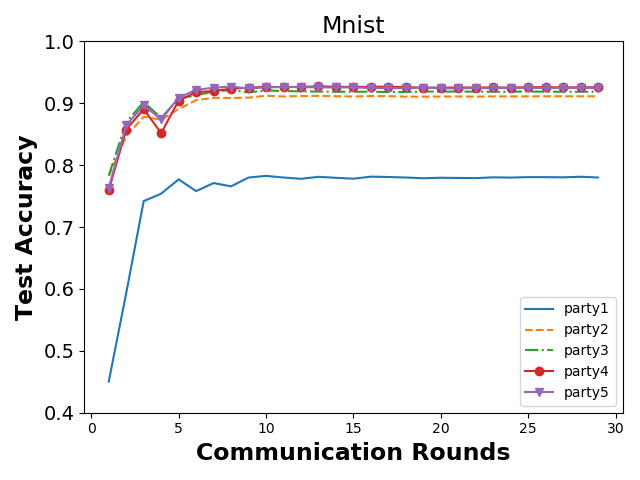}
        \end{subfigure}
        \begin{subfigure}[ht]{0.19\textwidth}
                \includegraphics[width=3.0cm,height=2.9cm]{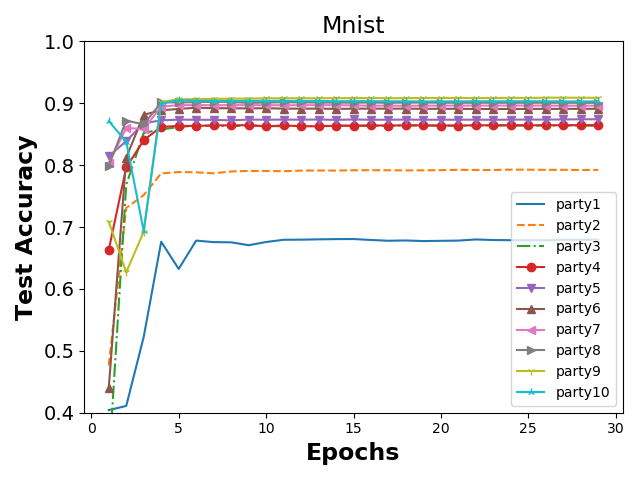}
        \end{subfigure}
        \begin{subfigure}[ht]{0.19\textwidth}
                \includegraphics[width=3.0cm,height=2.9cm]{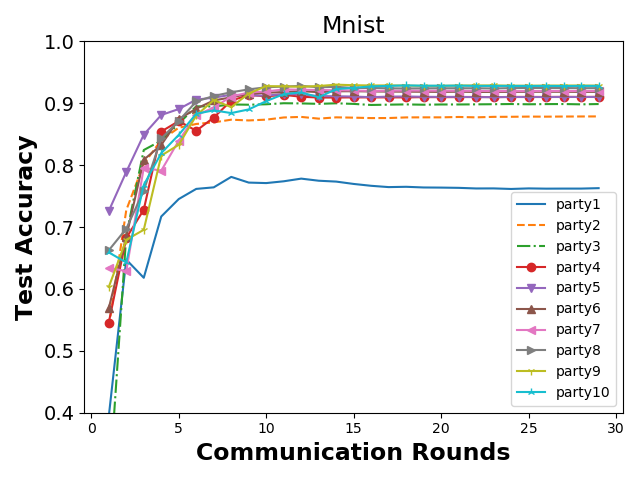}
        \end{subfigure}
         \begin{subfigure}[ht]{0.19\textwidth}
                 \includegraphics[width=3.0cm,height=2.9cm]{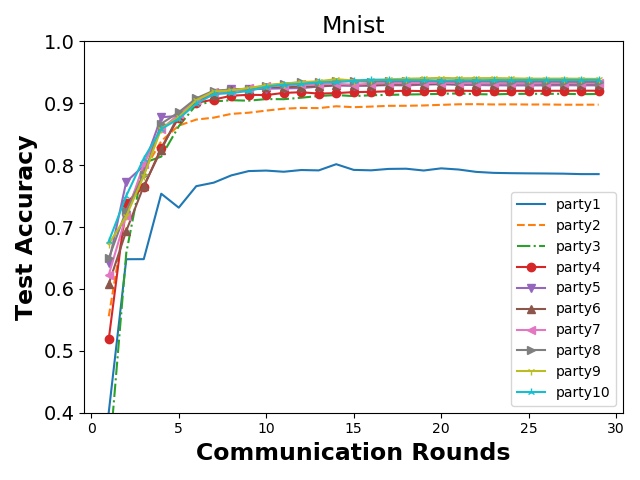}
        \end{subfigure}
        \begin{subfigure}[ht]{0.19\textwidth}
                \includegraphics[width=3.0cm,height=2.9cm]{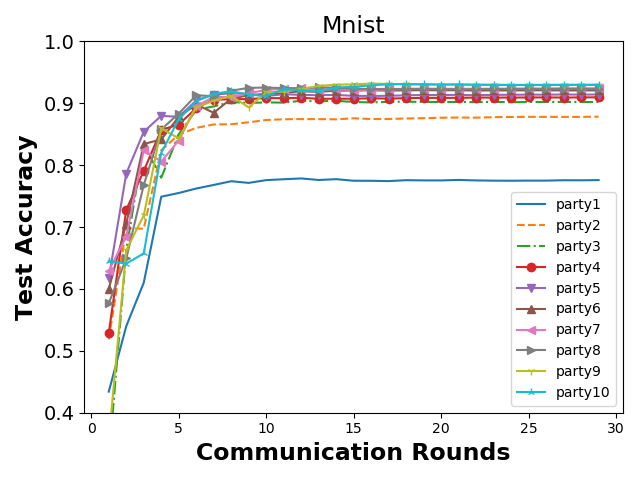}
        \end{subfigure}
         \begin{subfigure}[ht]{0.19\textwidth}
                 \includegraphics[width=3.0cm,height=2.9cm]{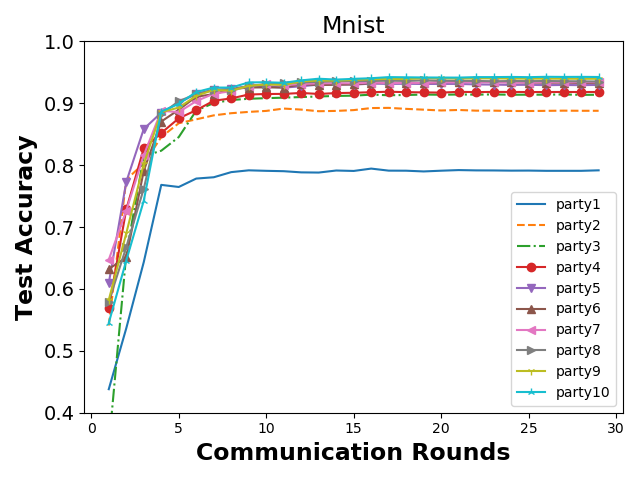}
        \end{subfigure}
         \begin{subfigure}[ht]{0.19\textwidth}
                \includegraphics[width=3.0cm,height=2.9cm]{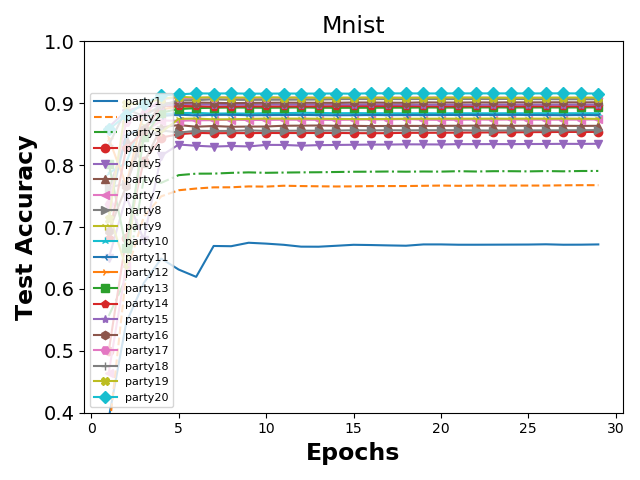}
        \end{subfigure}
        \begin{subfigure}[ht]{0.19\textwidth}
                \includegraphics[width=3.0cm,height=2.9cm]{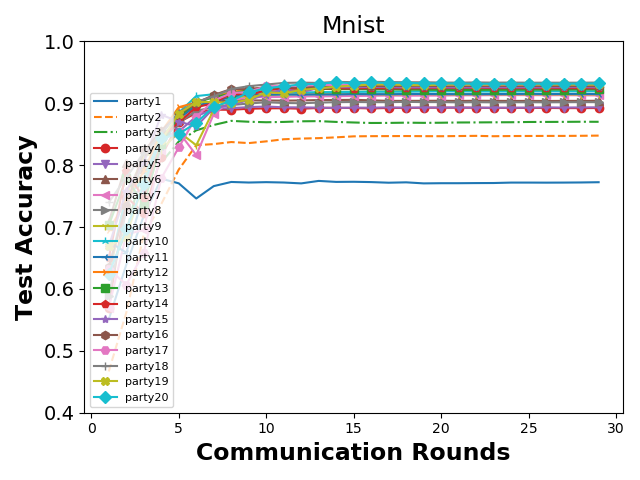}
        \end{subfigure}
         \begin{subfigure}[ht]{0.19\textwidth}
                 \includegraphics[width=3.0cm,height=2.9cm]{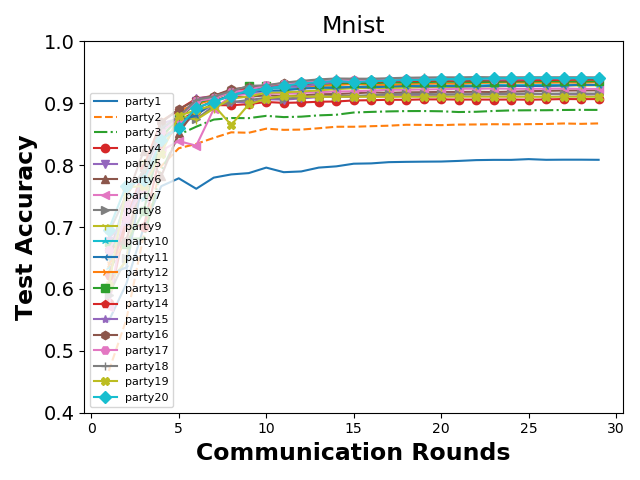}
        \end{subfigure}
        \begin{subfigure}[ht]{0.19\textwidth}
                \includegraphics[width=3.0cm,height=2.9cm]{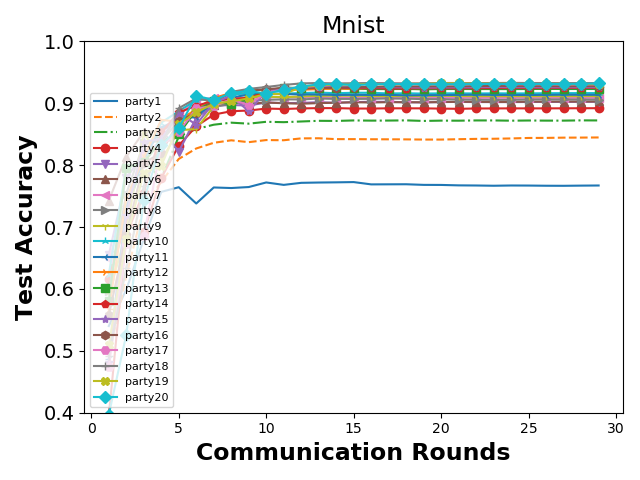}
        \end{subfigure}
         \begin{subfigure}[ht]{0.19\textwidth}
                 \includegraphics[width=3.0cm,height=2.9cm]{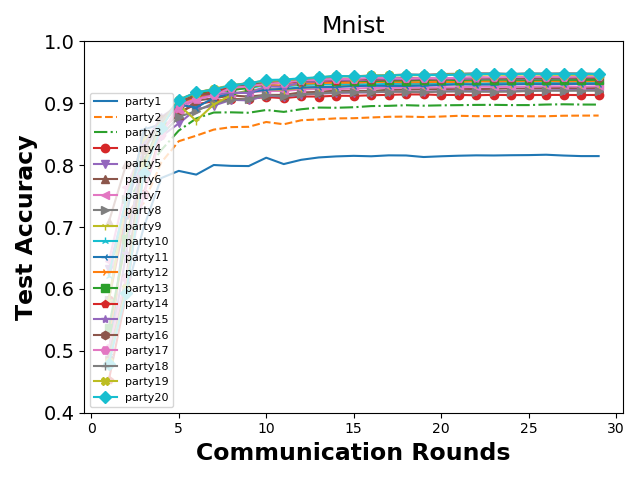}
        \end{subfigure}
         \caption{Individual convergence for MNIST using \textit{Standalone} framework and our CFFL. The 3 rows correspond to $\{5, 10, 20\}$ participants, the 5 columns correspond to \{Standalone, CFFL with $\theta_u=0.1$ and pretrain, CFFL with $\theta_u=1$\ and pretrain, CFFL with $\theta_u=0.1$ but without pretrain, CFFL with $\theta_u=1$\ but without pretrain\}.}
        % \caption{Individual convergence for MNIST using \textit{Standalone} framework (B=1, E=1, lr=0.001) and our CFFL (B=10, E=5, lr=0.1) with pretrain.}
\label{fig:mnist_mlp_convergence_pretrain1}
%}
\end{figure*}

\textbf{Individual model performance}. 
%\textbf{participant Convergence}. 
To examine the impact of our CFFL on individual convergence, Figure~\ref{fig:mnist_mlp_convergence_pretrain1} plots the test accuracy of each participant for the \textit{Standalone} framework and CFFL with upload rate of $\{0.1,1\}$ and with/without pretraining over MNIST across 30 (communication) rounds. It can be observed that our CFFL consistently delivers better accuracy than the standalone model of any participant.  Importantly, it confirms that our CFFL enforces the participants to converge to different local models, which are still better than their standalone models without collaboration, thereby offering fairness and utility as claimed. %We note that \emph{CFFL with pretraining} (2nd \& 3rd columns in Figure~\ref{fig:mnist_mlp_convergence_pretrain1}) may take slightly longer (less than 5 additional rounds) for all participants to converge than \emph{CFFL without pretraining} (4th \& 5th columns in Figure~\ref{fig:mnist_mlp_convergence_pretrain1}). Similar trends are observed for the Adult dataset from Figure~\ref{fig:adult_MLP_convergence}. 
We also observe slight fluctuations at the beginning of training. This can be attributed to the fact that participants are allocated with different aggregated updates from the server. As we can see from these figures, the convergence curves for CFFL (with pretrain) and CFFL (w/o pretrain) follow the similar trend, confirming that pretraining does not alter the overall convergence behaviour, but provides relatively better fairness in most cases.

\begin{figure*}[!htp]
%\resizebox{.5\totalheight}{!}{
\centering
        \begin{subfigure}[ht]{0.19\textwidth}
                \includegraphics[width=3.0cm,height=2.9cm]{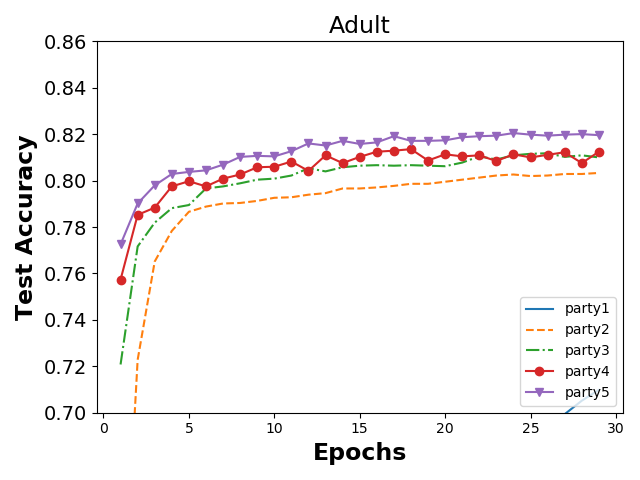}
        \end{subfigure}
        \begin{subfigure}[ht]{0.19\textwidth}
                \includegraphics[width=3.0cm,height=2.9cm]{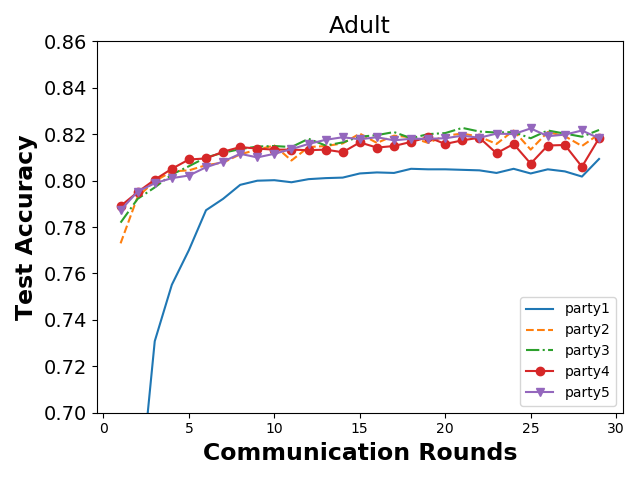}
        \end{subfigure}
         \begin{subfigure}[ht]{0.19\textwidth}
                \includegraphics[width=3.0cm,height=2.9cm]{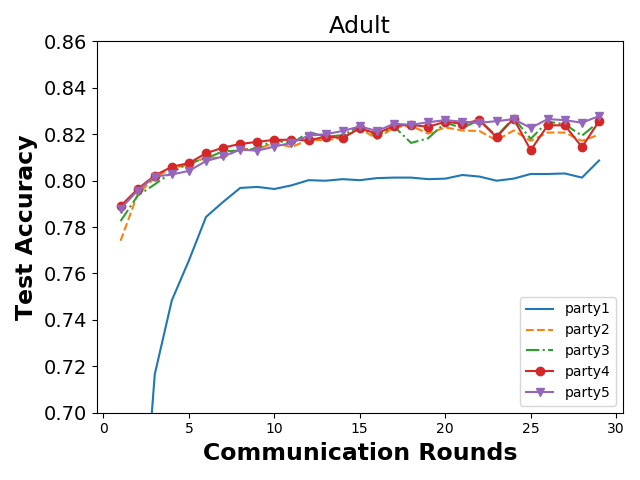}
        \end{subfigure}
        \begin{subfigure}[ht]{0.19\textwidth}
                \includegraphics[width=3.0cm,height=2.9cm]{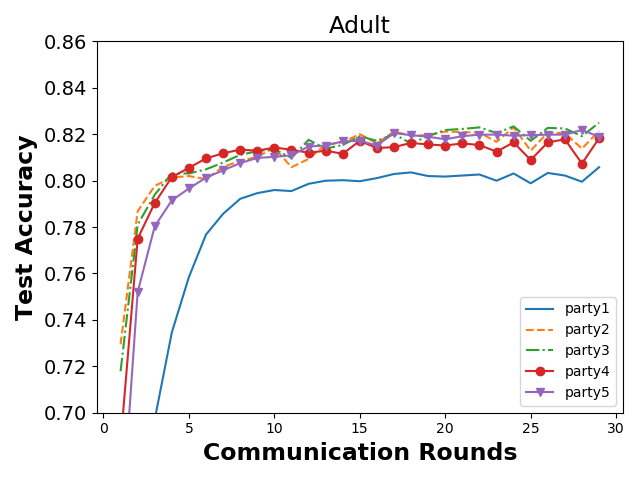}
        \end{subfigure}
         \begin{subfigure}[ht]{0.19\textwidth}
                \includegraphics[width=3.0cm,height=2.9cm]{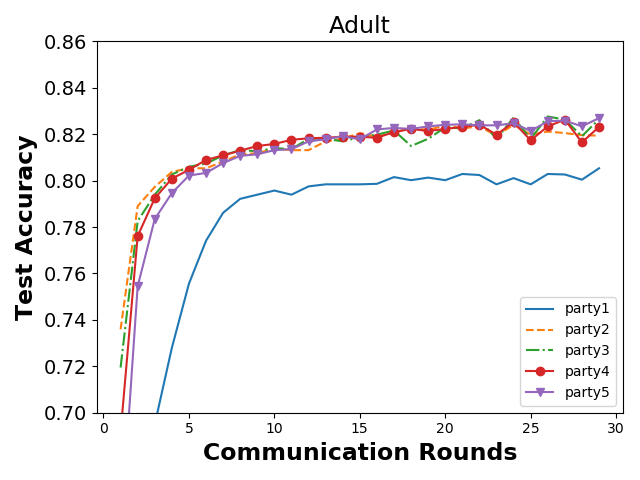}
        \end{subfigure}
        
        \begin{subfigure}[ht]{0.19\textwidth}
                \includegraphics[width=3.0cm,height=2.9cm]{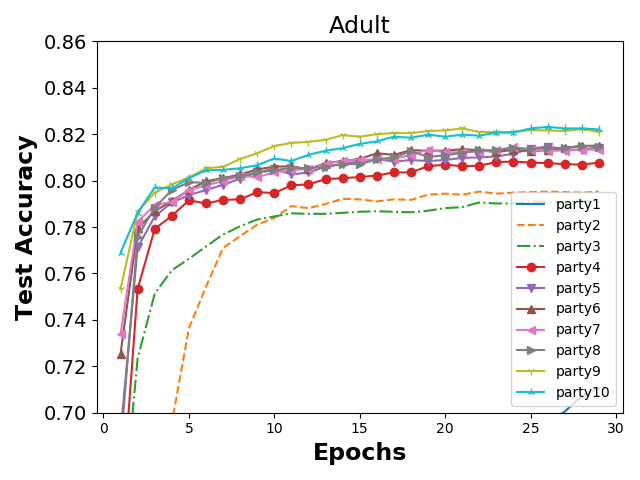}
        \end{subfigure}
        \begin{subfigure}[ht]{0.19\textwidth}
                \includegraphics[width=3.0cm,height=2.9cm]{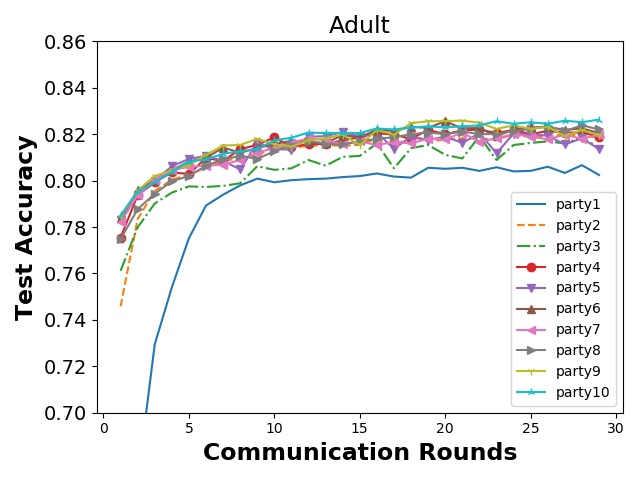}
        \end{subfigure}
         \begin{subfigure}[ht]{0.19\textwidth}
                \includegraphics[width=3.0cm,height=2.9cm]{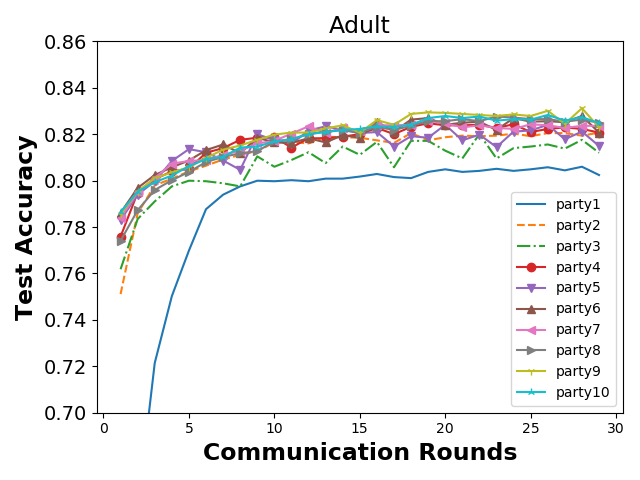}
        \end{subfigure}
        \begin{subfigure}[ht]{0.19\textwidth}
                \includegraphics[width=3.0cm,height=2.9cm]{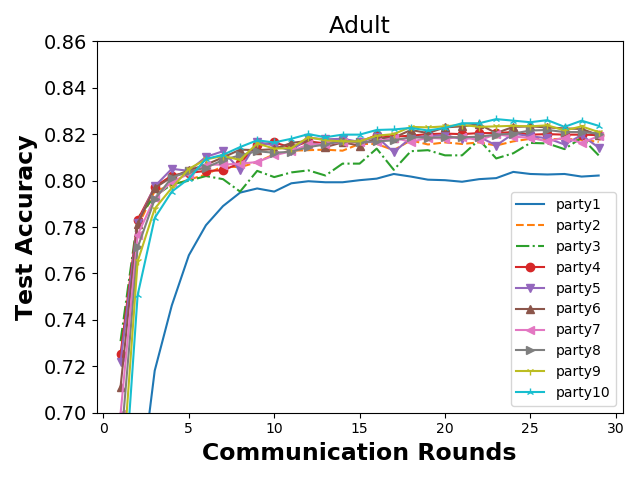}
        \end{subfigure}
         \begin{subfigure}[ht]{0.19\textwidth}
                \includegraphics[width=3.0cm,height=2.9cm]{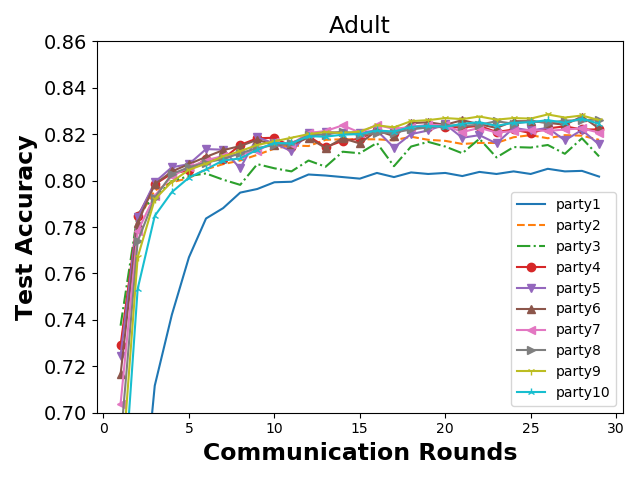}
        \end{subfigure}
         \begin{subfigure}[ht]{0.19\textwidth}
                \includegraphics[width=3.0cm,height=2.9cm]{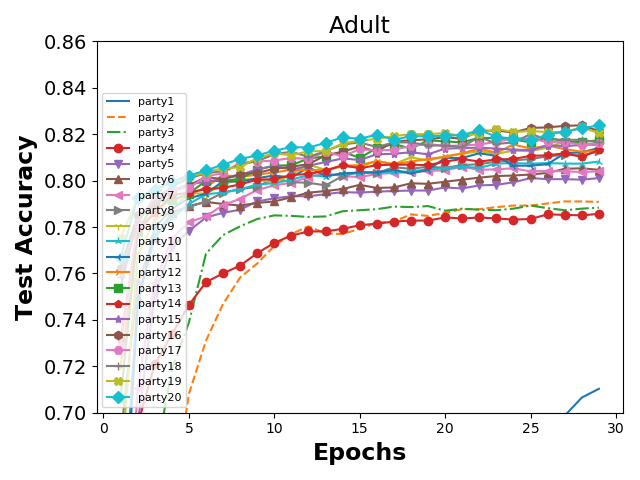}
        \end{subfigure}
        \begin{subfigure}[ht]{0.19\textwidth}
                \includegraphics[width=3.0cm,height=2.9cm]{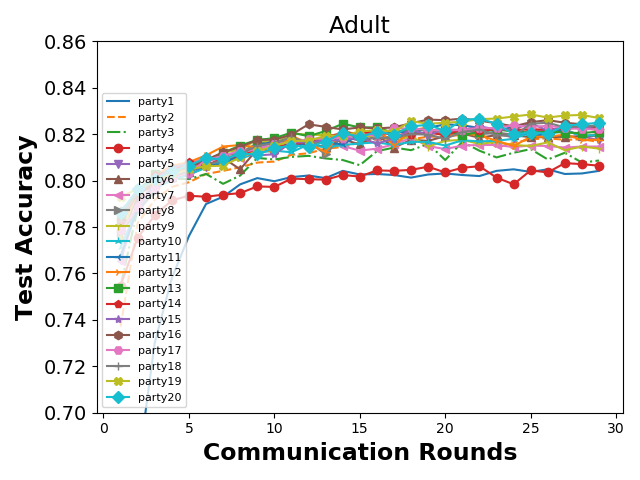}
        \end{subfigure}
         \begin{subfigure}[ht]{0.19\textwidth}
                \includegraphics[width=3.0cm,height=2.9cm]{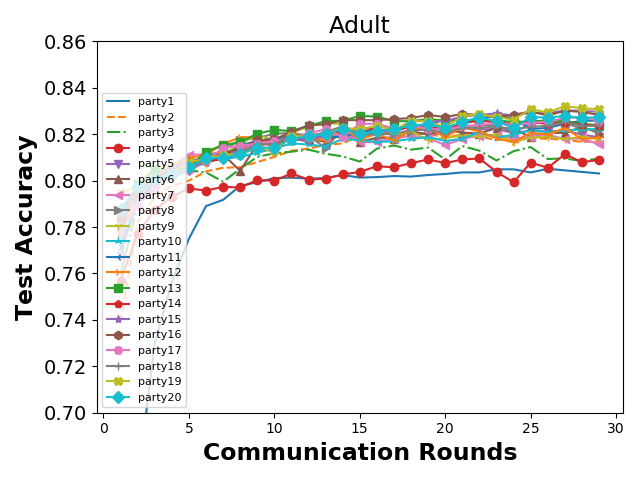}
        \end{subfigure}
        \begin{subfigure}[ht]{0.19\textwidth}
                \includegraphics[width=3.0cm,height=2.9cm]{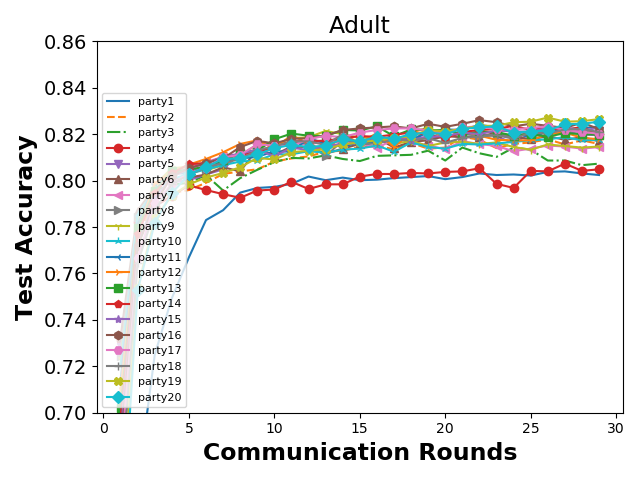}
        \end{subfigure}
         \begin{subfigure}[ht]{0.19\textwidth}
                \includegraphics[width=3.0cm,height=2.9cm]{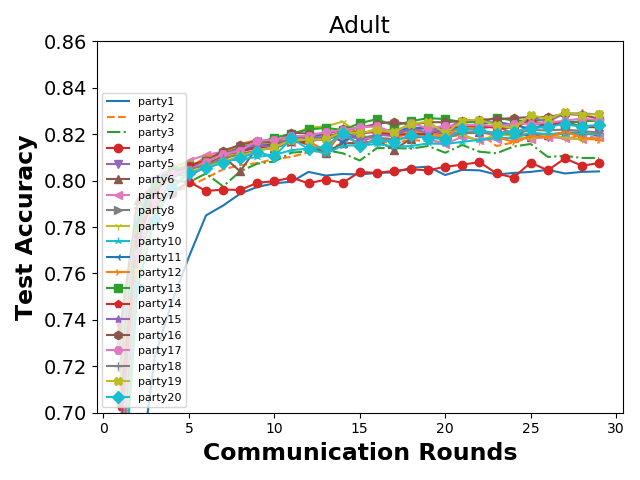}
        \end{subfigure}
         \caption{Individual convergence for Adult using \textit{Standalone} framework and our CFFL. The 3 rows correspond to $\{5, 10, 20\}$ participants, the 5 columns correspond to \{Standalone, CFFL with $\theta_u=0.1$ and pretrain, CFFL with $\theta_u=1$\ and pretrain, CFFL with $\theta_u=0.1$ but without pretrain, CFFL with $\theta_u=1$\ but without pretrain\}.}
\label{fig:adult_MLP_convergence}
%}
\end{figure*}

\begin{figure*}[!htp]
%\resizebox{.5\totalheight}{!}{
\centering
        \begin{subfigure}[ht]{0.19\textwidth}
                \includegraphics[width=3.0cm,height=2.9cm]{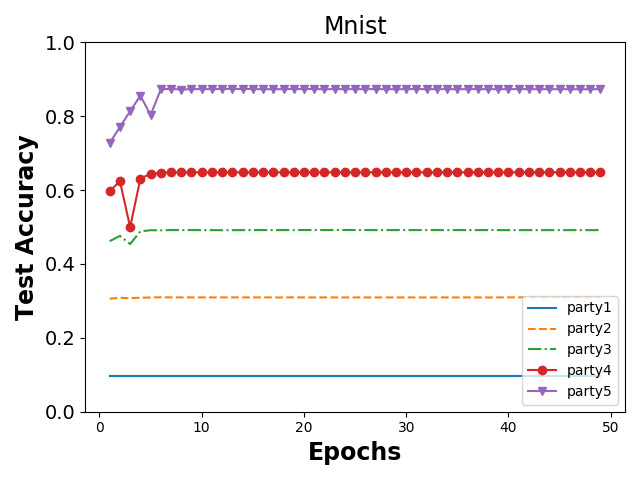}\label{fig:mnist_deep_p5e100_standalone}
                % \subcaption{Standalone (P5)}
        \end{subfigure}
        \begin{subfigure}[ht]{0.19\textwidth}
                \includegraphics[width=3.0cm,height=2.9cm]{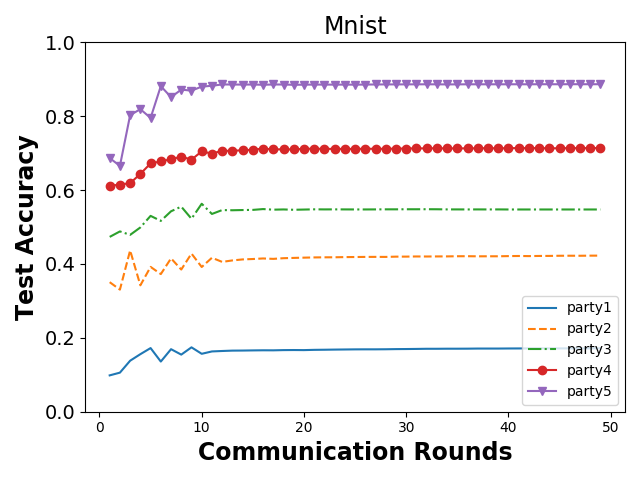}\label{fig:mnist_classimbalanced_cffl_p5_upload01_pretrain1}
                % \subcaption{CFFL (P5, $\theta_u=0.1$)}
        \end{subfigure}
         \begin{subfigure}[ht]{0.19\textwidth}
                \includegraphics[width=3.0cm,height=2.9cm]{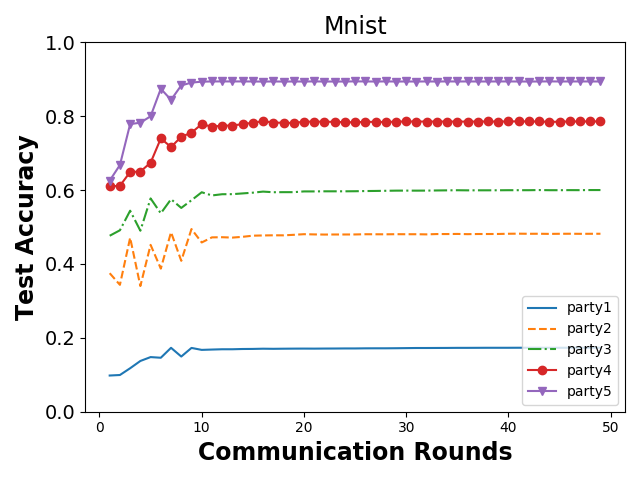}\label{fig:mnist_classimbalanced_cffl_p5_upload1_pretrain1}
                % \subcaption{CFFL (P5, $\theta_u=1$)}
        \end{subfigure}
        \begin{subfigure}[ht]{0.19\textwidth}
                \includegraphics[width=3.0cm,height=2.9cm]{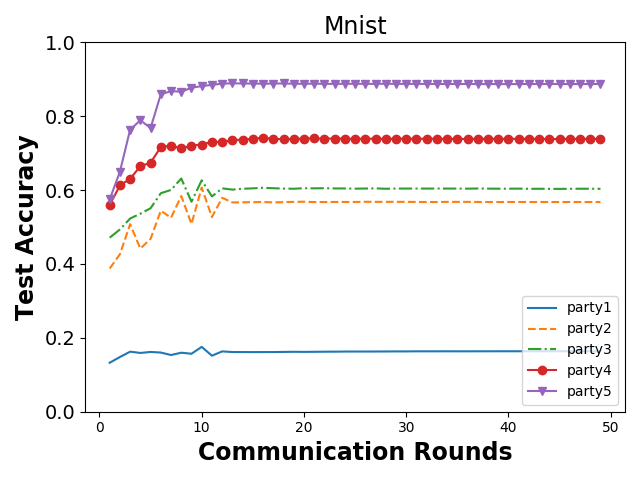}\label{fig:mnist_classimbalanced_cffl_p5_upload01_pretrain0}
                % \subcaption{CFFL (P5, $\theta_u=0.1$)}
        \end{subfigure}
         \begin{subfigure}[ht]{0.19\textwidth}
                \includegraphics[width=3.0cm,height=2.9cm]{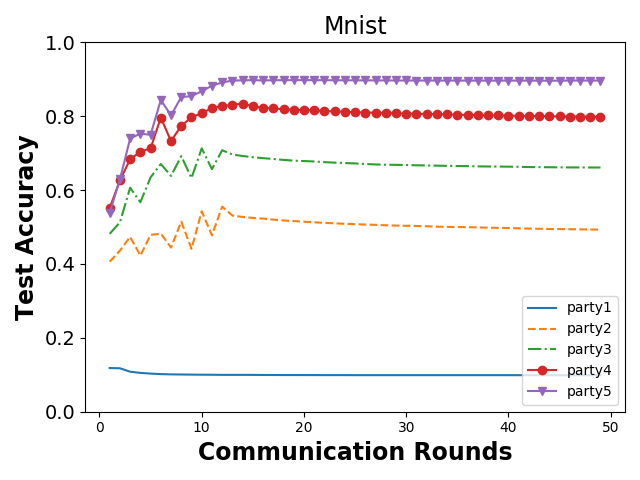}\label{fig:mnist_classimbalanced_cffl_p5_upload1_pretrain0}
                % \subcaption{CFFL (P5, $\theta_u=1$)}
        \end{subfigure}
        
        \begin{subfigure}[ht]{0.19\textwidth}
                \includegraphics[width=3.0cm,height=2.9cm]{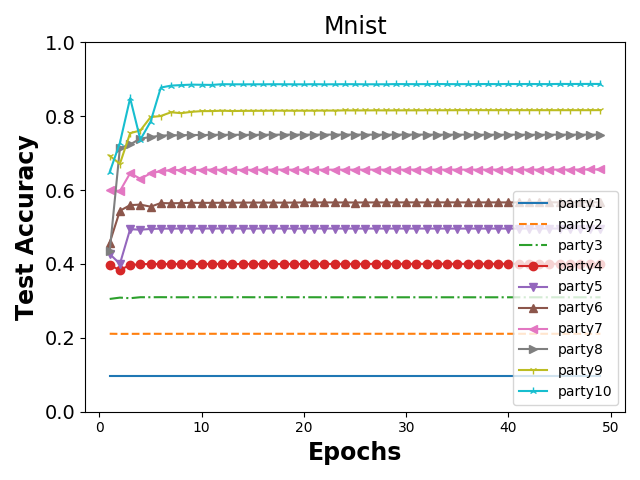}\label{fig:mnist_deep_p10e100_standalone}
                % \subcaption{Standalone (p10)}
        \end{subfigure}
        \begin{subfigure}[ht]{0.19\textwidth}
                \includegraphics[width=3.0cm,height=2.9cm]{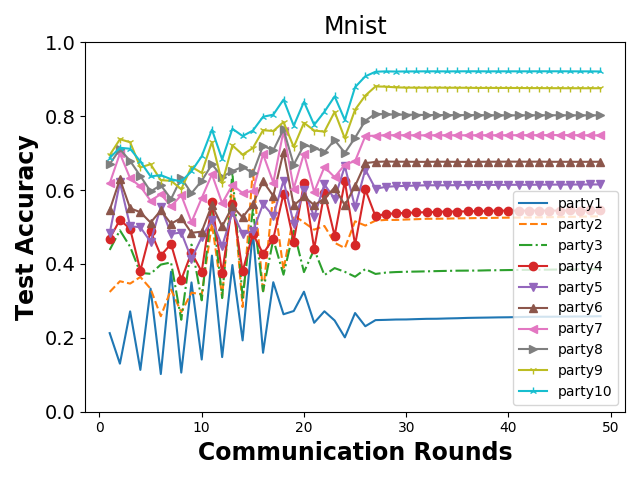}\label{fig:mnist_classimbalanced_cffl_p10_upload01_pretrain1}
                % \subcaption{CFFL (p10, $\theta_u=0.1$)}
        \end{subfigure}
         \begin{subfigure}[ht]{0.19\textwidth}
                \includegraphics[width=3.0cm,height=2.9cm]{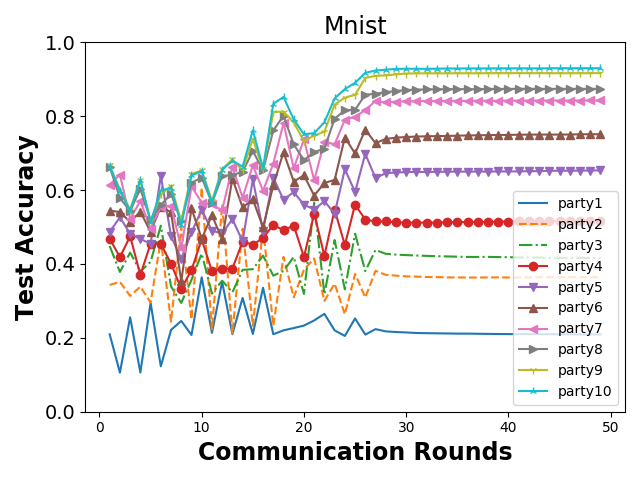}\label{fig:mnist_classimbalanced_cffl_p10_upload1_pretrain1}
                % \subcaption{CFFL (p10, $\theta_u=1$)}
        \end{subfigure}
        \begin{subfigure}[ht]{0.19\textwidth}
                \includegraphics[width=3.0cm,height=2.9cm]{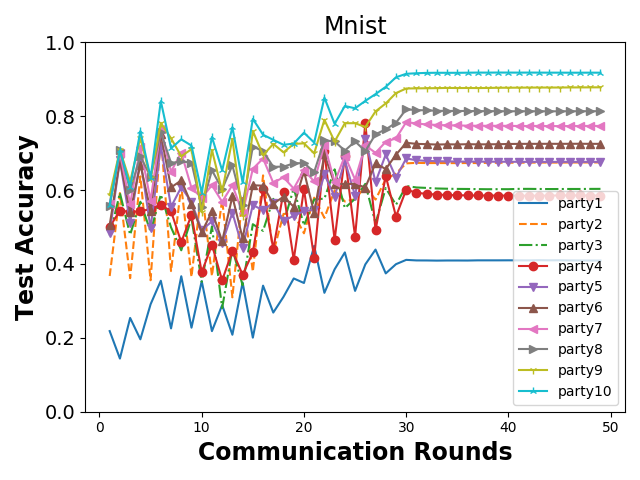}\label{fig:mnist_classimbalanced_cffl_p10_upload01_pretrain0}
                % \subcaption{CFFL (p10, $\theta_u=0.1$)}
        \end{subfigure}
         \begin{subfigure}[ht]{0.19\textwidth}
                \includegraphics[width=3.0cm,height=2.9cm]{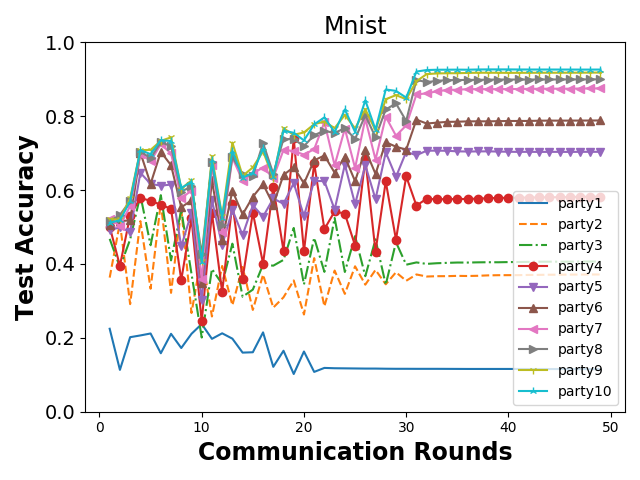}\label{fig:mnist_classimbalanced_cffl_p10_upload1_pretrain0}
                % \subcaption{CFFL (p10, $\theta_u=1$)}
        \end{subfigure}
        
        \begin{subfigure}[ht]{0.19\textwidth}
                \includegraphics[width=3.0cm,height=2.9cm]{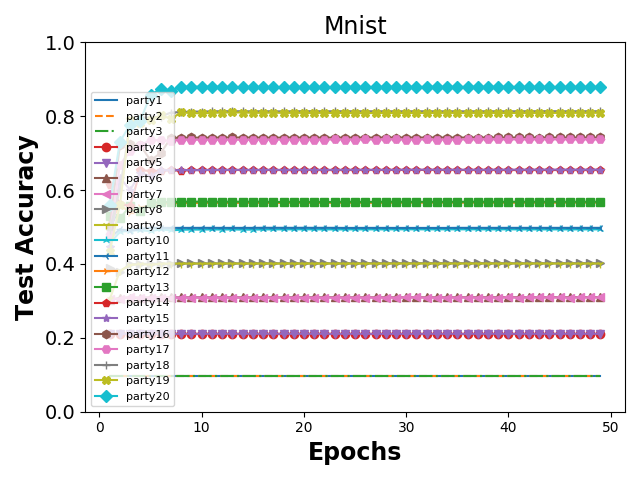}\label{fig:mnist_deep_p20e100_standalone}
                % \subcaption{Standalone (p20)}
        \end{subfigure}
        \begin{subfigure}[ht]{0.19\textwidth}
                \includegraphics[width=3.0cm,height=2.9cm]{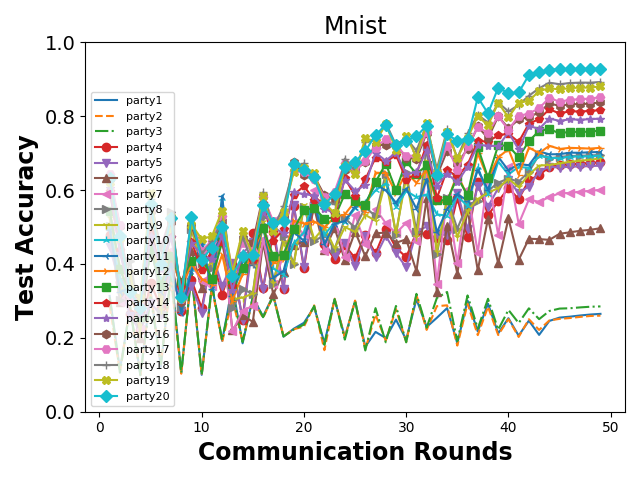}\label{fig:mnist_classimbalanced_cffl_p20_upload01_pretrain1}
                % \subcaption{CFFL (p20, $\theta_u=0.1$)}
        \end{subfigure}
         \begin{subfigure}[ht]{0.19\textwidth}
                \includegraphics[width=3.0cm,height=2.9cm]{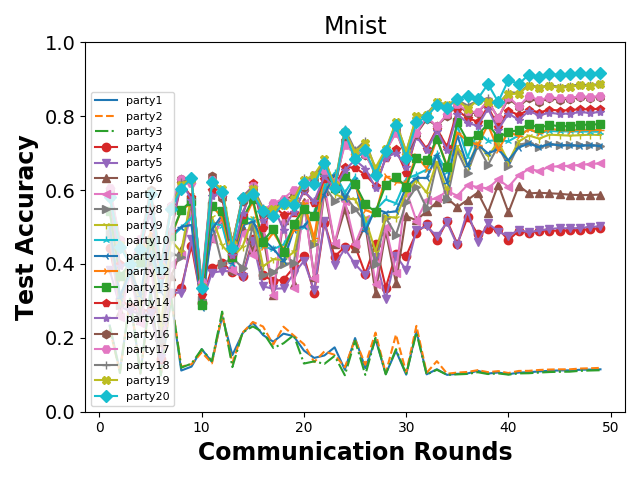}\label{fig:mnist_classimbalanced_cffl_p20_upload1_pretrain1}
                % \subcaption{CFFL (p20, $\theta_u=1$)}
        \end{subfigure}
        \begin{subfigure}[ht]{0.19\textwidth}
                \includegraphics[width=3.0cm,height=2.9cm]{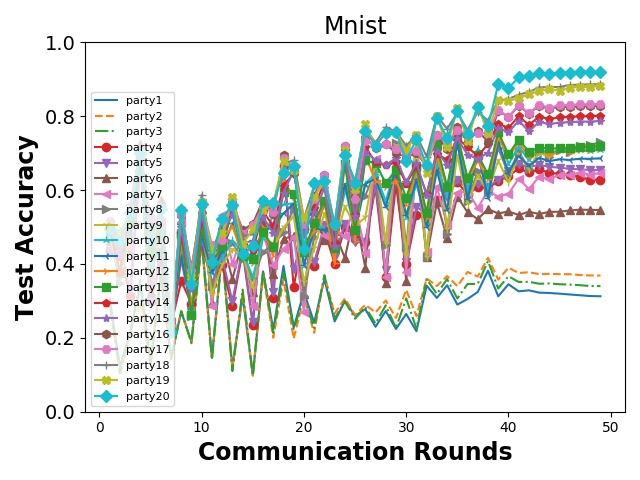}\label{fig:mnist_classimbalanced_cffl_p20_upload01_pretrain0}
                % \subcaption{CFFL (p20, $\theta_u=0.1$)}
        \end{subfigure}
         \begin{subfigure}[ht]{0.19\textwidth}
                \includegraphics[width=3.0cm,height=2.9cm]{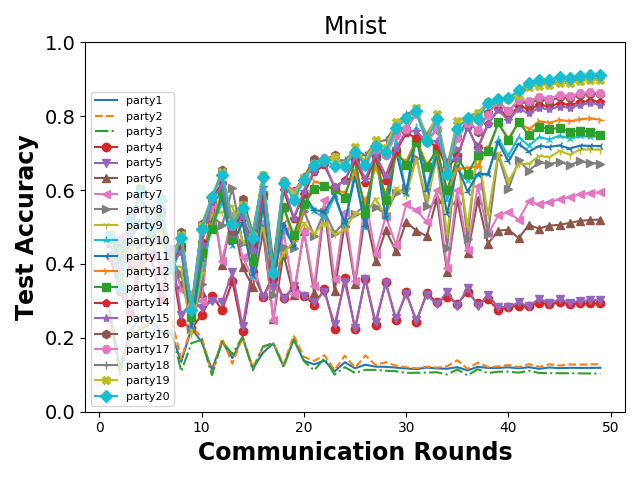}\label{fig:mnist_classimbalanced_cffl_p20_upload1_pretrain0}
                % \subcaption{CFFL (p20, $\theta_u=1$)}
        \end{subfigure}
         \caption{Individual model accuracy for MNIST class imbalanced scenario, where classes are distributed in a linspace manner (for example, participant-$\{1,2,3,4,5\}$ own \{1,3,5,7,10\} classes respectively). 5 columns correspond to \{Standalone, CFFL $\theta_u=0.1$ w pretrain, CFFL $\theta_u=1$\ w pretrain, CFFL $\theta_u=0.1$ w/o pretrain, CFFL $\theta_u=1$\ w/o pretrain\}.}
\label{fig:mnist_classimbalanced_acc}
\end{figure*}

For imbalanced class numbers, Figure~\ref{fig:mnist_classimbalanced_acc} shows individual model accuracy in the \emph{Standalone} framework and our CFFL. We see that all participants achieve higher accuracies in CFFL than their standalone counterparts. Similar to the scenario of imbalanced data size, all participants converge to different final models, but with more obvious accuracy gaps, resulting in higher fairness. Moreover, it takes longer for participants to converge when there are more participants in the system.

%%%%%%%%%%%%%%%%%%%%%%%%%%%%%%%%%%%%%%%%
\section{Discussions}
\label{sec:Discussion}
\textbf{Robustness to Free-riders.}  
In an FL system, there may exist free-riders who aim to benefit from the global model, without really contributing. Typically, free-riders may pretend to be contributing by %generating and 
uploading random or noisy updates. In standard FL systems, there is no specific safeguard against this, so even free-riders can enjoy the system's global model at virtually no cost. Conversely, CFFL can automatically identify and isolate free-riders. This is because the empirical utility (on the validation set) of the random or noisy gradients is generally low. As collaborative training proceeds, the free-riders will receive gradually lower reputations, and eventually be isolated from the system when their reputations fall below the reputation threshold. Through our additional experiments (including 1 free rider who always uploads random values as gradients), we observe that our CFFL can always identify and isolate the free rider at the early stages of collaboration, without affecting both accuracy and convergence.

\textbf{Choice of Reputation Threshold}. Using a reputation threshold $c_{th}$ allows the server to enforce a lower bound on the reputation. This can be used to detect and isolate the free-rider in the system. A key challenge lies in the selection of an appropriate threshold, as fairness and accuracy may be inversely affected. For example, too small $c_{th}$ might allow low-contribution participant to sneak into the federated system without being detected and isolated. On the contrary, too large $c_{th}$ might isolate too many participants in the system. In our experiments, we empirically find suitable values for different scenarios.%, \ie $c_{th}=\frac{1}{|R|}*\frac{1}{3}$ for imbalanced data size, and $c_{th}=\frac{1}{|R|}*\frac{1}{6}$ for imbalanced class numbers.

\section{Conclusion and Future Work}
\label{sec:Conclusion}
This work initiates the research %problem of 
on collaborative fairness in federated learning (FL), and modifies FL to enforce participants to converge to different models. A novel collaborative fair federated learning framework named CFFL is proposed. Based on empirical individual model performance on a validation set, a reputation mechanism is introduced to mediate participant rewards across communication rounds. Experimental results demonstrate that CFFL achieves comparable accuracy to two \textit{Distributed} frameworks, and always achieves better accuracy than the \textit{Standalone} framework, confirming the effectiveness of CFFL in terms of both \emph{fairness} and \emph{utility}. A number of avenues for further work are appealing. In particular, we would like to study how to quantify fairness in more complex settings, % involving other data distributions, 
and %other types of learning tasks 
apply our framework to various domains, such as financial, biomedical, speech, NLP, etc. Furthermore, we would like to systematically integrate robustness with fairness. %investigate the robustness of CFFL under different assumptions about the participants, such as adversaries. 
It is expected that our system can find wide applications in real world.

% \section*{Acknowledgments}
% This research is supported by the 

\bibliographystyle{IEEEbib}
\bibliography{biblio.bib}

\end{document}